\documentclass[sn-nature]{sn-jnl}

\usepackage{graphicx}%
\usepackage{multirow}%
\usepackage{amsmath,amssymb,amsfonts}%
\usepackage{amsthm}%
\usepackage{mathrsfs}%
\usepackage[title]{appendix}%
\usepackage{xcolor}%
\usepackage{textcomp}%
\usepackage{manyfoot}%
\usepackage{booktabs}%
\usepackage{algorithm}%
\usepackage{algorithmicx}%
\usepackage{algpseudocode}%
\usepackage{listings}%
\usepackage{fancyhdr}%

\begin{document}

\title[Article Title]{\textbf{A Foundational Framework and Methodology for Personalized Early and Timely Diagnosis}}

\author[1,2]{\fnm{Tim} \sur{Schubert}}

\author[3,4,5]{\fnm{Richard W} \sur{Peck}}

\author[5]{\fnm{Alexander} \sur{Gimson}}

\author[6]{\fnm{Camelia} \sur{Davtyan}}

\author*[1,5]{\fnm{Mihaela} \sur{van der Schaar}}\email{mv472@damtp.cam.ac.uk}

\affil[1]{\orgdiv{Department of Applied Mathematics and Theoretical Physics}, \orgname{University of Cambridge}, \orgaddress{\city{Cambridge}, \country{UK}}}

\affil[2]{\orgdiv{Institute of Human Genetics}, \orgname{Heidelberg University}, \orgaddress{\city{Heidelberg}, \country{Germany}}}

\affil[3]{\orgdiv{Department of Pharmacology and Therapeutics}, \orgname{University of Liverpool}, \orgaddress{\city{Liverpool}, \country{UK}}}

\affil[4]{\orgdiv{Pharma Research \& Early Development (pRED)}, \orgname{Roche Innovation Center}, \orgaddress{\city{Basel}, \country{Switzerland}}}

\affil[5]{\orgdiv{The Cambridge Centre for AI in Medicine}, \orgaddress{\city{Cambridge}, \country{UK}}}

\affil[6]{\orgdiv{Department of Internal Medicine}, \orgname{University of California Los Angeles}, \orgaddress{\city{Los Angeles}, \country{USA}}}

\abstract{Early diagnosis of diseases holds the potential for deep transformation in healthcare by enabling better treatment options, improving long-term survival and quality of life, and reducing overall cost. With the advent of electronic health records, increasing amounts of patient-collected data, and advances in diagnostic tests as well as in machine learning and statistics, early or timely diagnosis seems within reach. Early diagnosis research often focuses on novel technologies and risk stratification while neglecting the potential for optimizing individual diagnostic paths. To enable personalized early and timely diagnosis, a foundational framework is needed that delineates the diagnosis process and systematically identifies the time-dependent value of various diagnostic tests for an individual patient given their unique characteristics and history.
\\

Decision-theoretic models have been proposed as a potential solution to conceptualize the diagnosis process but 1) they are not individualized to the specific characteristics of the patient, and 2) they often neglect the fact that costs and benefits of diagnostic decisions are not known and need to be estimated from data.
\\

In this paper, we propose the first foundational framework for early and timely diagnosis. 

Our framework builds on decision-theoretic approaches to systematically outline the diagnosis process and integrates machine learning and statistical methodology for estimating the optimal personalized diagnostic path. Moreover, to describe the proposed framework as well as possibly other frameworks, we provide a set of essential definitions.
\\

The development of a foundational framework is necessary for several reasons: 

1) formalism provides clarity for the development and implementation of methods and tools to support diagnostic decision-making;

2) observed information from present and past can be complemented with estimates of the future patient trajectory;

3) the net benefit of counterfactual diagnostic paths and the associated uncertainties can be modeled for individual patients to inform shared decision-making;

4) concepts such as 'early' and 'timely' diagnosis can be clearly defined, differentiated and understood; 

5) a systematic mechanism emerges for assessing the value of technologies, including new and existing diagnostic tests in terms of their impact on personalized early diagnosis, resulting health outcomes and incurred costs for individuals, populations and healthcare systems.
\\

Finally, we hope that this foundational framework will unlock the long-awaited potential of timely diagnosis and intervention, leading to improved outcomes for patients and higher cost-effectiveness for healthcare systems.
}

\keywords{early diagnosis, timely diagnosis, framework, medical decision-making, decision support, machine learning, decision theory, precision medicine, healthcare, medicine, statistics}



\maketitle

\section{Introduction}\label{sec1}

Diagnosis is a cornerstone of healthcare, acting as a working hypothesis to draw conclusions about disease trajectories and to dictate the use of diagnostic tools and potential therapies.
Numerous studies have demonstrated the benefits of earlier diagnosis and treatment of disease which promise substantial advantages for patient outcomes and the healthcare system by enabling less invasive treatment options, long-term survival, improved quality of life, and reduced overall cost.\cite{Neal2015-ob}\cite{Laudicella2016-ft}\cite{Cobo-Calvo2023-lt}\cite{Aletaha2018-wj}\cite{noauthor_2021-uq}\cite{Sennfalt2023-fj} The need for early intervention is evident across most medical domains. For instance, every four weeks of delay in starting adjuvant systemic treatment for colorectal cancer is associated with an average 13\% increase in risk of death\cite{Hanna2020-zc} and early diagnosis of heart failure can delay disease progression and late-stage complications\cite{De_Couto2010-mv}.

\medskip
Early detection and diagnosis are emerging as a priority for organizations across the globe\cite{Crosby2020-yw} leading to ambitious initiatives and new diagnostic tools.\cite{Nordberg2015-tj}\cite{Liu2020-cl} For instance, in cancer research, efforts are made to better understand cancer biology, find biomarkers, and develop new technologies and risk stratification models.\cite{Crosby2022-jo} However, the need to optimize the diagnostic process itself is neglected. Improving this process is crucial, as this has the potential to yield better health outcomes for patients while also reducing healthcare costs. Many diagnostic tests both novel (such as the Galleri test\cite{Liu2020-cl}) and conventional (such as MRIs) come with significant costs, including those associated with overdiagnosis and unnecessary treatment, and their utility depends on the patient, disease and timing.\cite{Doubeni2023-sm}\cite{US_Preventive_Services_Task_Force2017-es} Particularly with the expanding array of available tools, it becomes essential to understand the aim and process of medical diagnosis at a fundamental level to inform choices among potential diagnostic actions and optimize diagnostic paths for individual patients.

\medskip
Policy documents often present the benefits of ‘early’ diagnosis as axiomatic.\cite{Dhedhi2014-dq} However, identifying a disease early, perhaps with extensive testing, when there is no additional benefit from earlier diagnosis is not necessarily in a patient’s best interest. Instead, clinicians and patients value the timeliness of a diagnosis\cite{Dhedhi2014-dq} which can be loosely viewed as diagnosing the “right person, to the right extent, at the right time, [...] in the right way”\cite{Aristotle2016-lw}. Achieving timely diagnosis for a patient requires personalization because what is timely for one patient may not be so for another. Variation in disease pathogenesis, presentation, progression, and in the effects of treatment between patients may mean that one patient would benefit from diagnosis at a different time from another patient. The greatest benefit from diagnosis for each patient will come from taking this variability into account which presents additional challenges.

\medskip
Decision theory has been used to analyze medical decision processes and aid clinicians in making complex decisions under uncertainty by assigning probabilities of outcomes and risks for each potential action.\cite{Cohen1996-ic} However, such approaches have not addressed two important challenges: 1) the need for personalization to individual patients, and 2) the lack of rationales for measuring personalized costs and benefits of diagnostic decisions against a goal and estimating them from data. We believe that to solve the challenge of early and timely diagnosis, it is imperative to develop a formal framework for medical diagnosis optimized to individual patients.

\medskip
A formal framework will solve several challenges: 

1) providing formalism to link the requirement for timely diagnosis with the development and implementation of the necessary methodology;

2) estimating the future patient trajectory from observed information;

3) modeling counterfactual diagnostic paths and the associated net benefit and uncertainties for individual patients to inform shared decision-making;

4) differentiating between ‘early’ and ‘timely’ diagnosis; and 

5) evaluating the benefits of technologies, including new and existing diagnostic tests, statistical methods, and machine learning solutions in terms of their impact on personalized timely diagnosis, resulting health outcomes and personal as well as systemic costs.
\medskip

This paper presents a foundational framework for personalized early and timely diagnosis, providing supporting definitions for the goal(s) and key aspects of the framework. The definitions themselves do not constitute the framework; rather, the framework utilizes these definitions to path the way for timely diagnosis. Clear definitions are essential to ensure agreement (or clarity on areas of disagreement), a precise understanding of the goal(s) of the framework, transparency, understandability, and reproducibility. Moreover, alternative frameworks can be developed based on the same, agreed-upon definitions and compared to the proposed framework.

\medskip
We outline challenges and solutions associated with the framework and explain how the advent of medical big data through electronic health records (EHR) and patient-collected data, combined with methodological advances in machine learning and statistics, enables the estimation of future patient trajectories and counterfactual costs and benefits of various possible diagnostic paths for individual patients. With these developments, implementation of such a framework for timely diagnosis becomes possible for the first time.

\medskip
Ultimately, our vision is that once implemented in clinical practice, this framework will empower clinicians and patients to better understand the strengths and weaknesses of their options during medical diagnosis and support them in making shared decisions.

\section{\textbf{The Foundational Framework for Personalized Early and Timely Diagnosis}}\label{sec2}

The foundational framework positions the patient and clinician at the center of the diagnosis process and provides actionable insights to facilitate the best patient outcome. Its implementation will require the synergy of multidisciplinary expertise from clinical medicine, ethics, and health economics to statistics and machine learning. 

\subsection{Definition of elements of the diagnostic process}\label{subsec1}

We draw from fields such as decision theory to provide precise terminology, allowing us to systematically formalize the intricacies and goals of medical diagnosis. In \autoref{table1}, we define the elements of a diagnostic process that are crucial for comprehending this framework and any potential frameworks proposed in the future.

\begin{table*}[h]
    \centering
    
    \begin{tabular}{@{}lp{0.6\linewidth}@{}}
    \toprule
    Term & Definition \\
    \midrule

    \medskip
    \bfseries Diagnosis & a label (health or disease) applied to a person at a specific time\\ 
    \medskip
    \bfseries Differential diagnoses & the set of possible diagnoses for a patient\\ 
        \addlinespace[0.5pt]
    \bigskip
    \bfseries Disease trajectory & the course of a disease over time\\ 
    
    \bigskip
    \bfseries Diagnostic uncertainty & manifestation of information deficiency\cite{Klir2006-ia} expressed with probabilities\\ 
    
    \bigskip
    \bfseries Diagnostic information & available information with the capacity to reduce diagnostic uncertainty\cite{Klir2006-ia} and ambiguity\\ 
    
    \bigskip
    \bfseries Diagnostic action & action (including no action) at a specific time to gather additional information in order to facilitate diagnosis or make diagnostic decisions\\ 
   
    \bigskip
    \bfseries Diagnostic path & series of diagnostic actions\cite{Rao2017-mc} taken at specific times to arrive at a diagnosis\\ 
    
    \bigskip
    \bfseries Diagnostic decision & choice of the subsequent diagnostic path in the face of bounded diagnostic uncertainty\\ 
   
    \bigskip
    \bfseries Diagnostic considerations & the time- and context-dependent benefits and costs to be considered along a diagnostic path\\ 
    
    \bigskip
    \bfseries Diagnostic net benefit & the sum of diagnostic considerations\\ 
    
    \bigskip
    \bfseries Early diagnosis & diagnosis following the diagnostic path which results in a diagnosis at the earliest possible time\cite{World_Health_Organization2017-jh} \\
    
    \bigskip
    \bfseries Timely diagnosis & diagnosis following the diagnostic path with the highest diagnostic net benefit\\
    \bottomrule
    \end{tabular}
    
    \bigskip
    \caption{Definitions describing the diagnosis process.}\label{table1}%
\end{table*}

\subsection{Description of the foundational framework}\label{subsec2}

\begin{figure}
    \centering
    \includegraphics[width=0.4\textwidth]{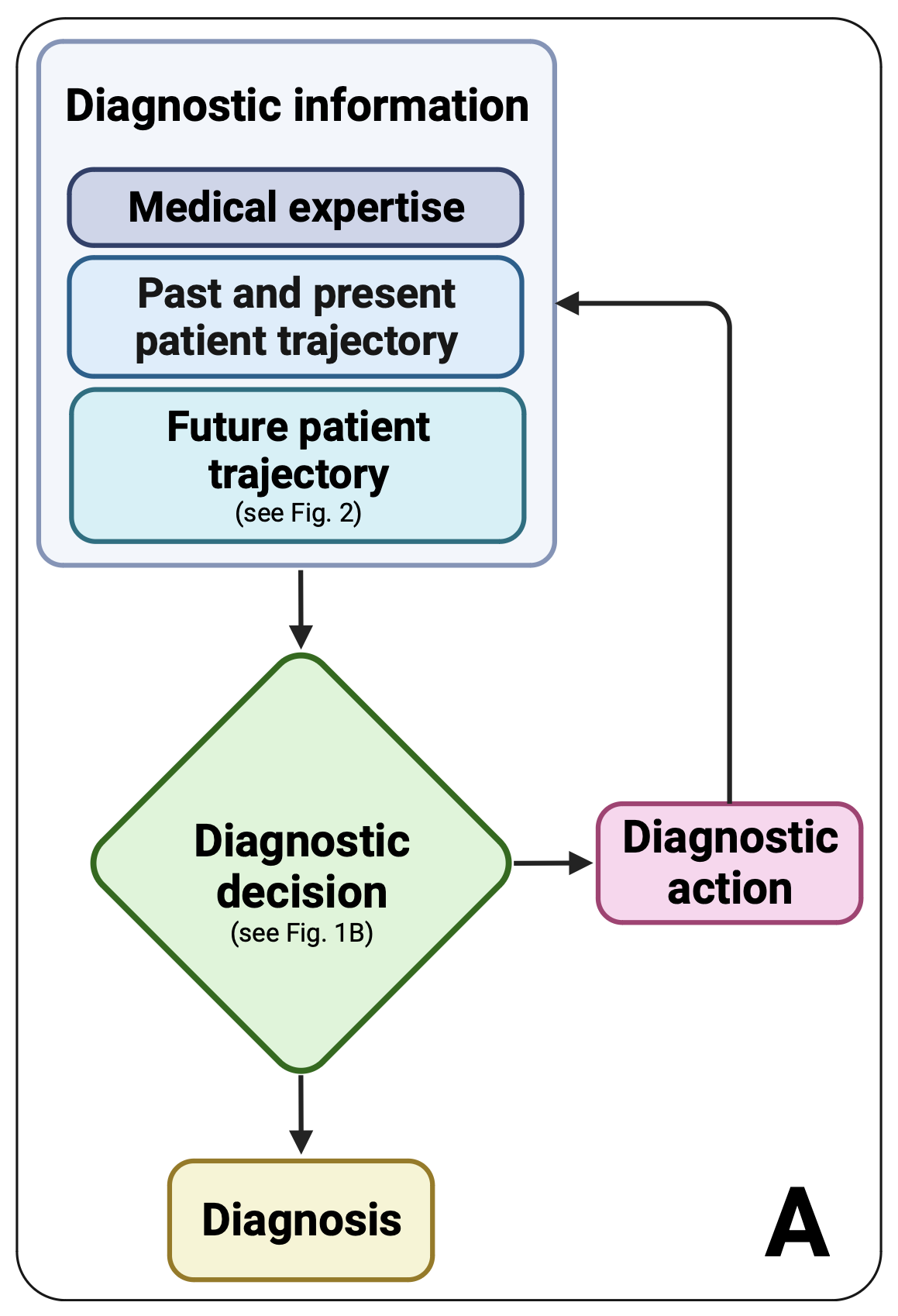}
    \includegraphics[width=0.9\textwidth]{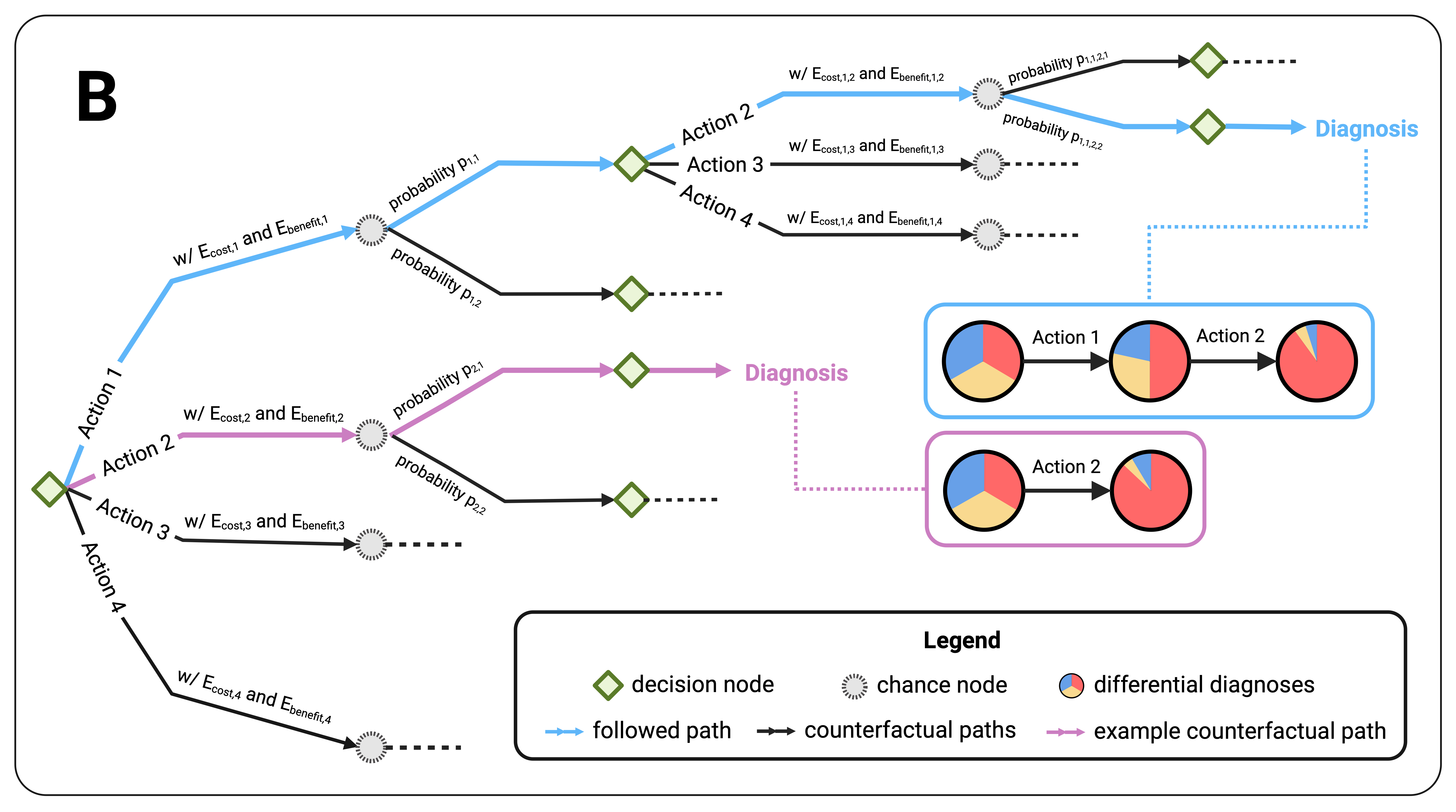}
    \caption{\textbf{Framework for personalized diagnosis.} \textbf{A) Diagnostic flowchart}. Diagnostic information is unique for each patient and entails medical expertise, past and present patient trajectory as well as estimates of the future patient trajectory. To enable timely diagnosis these estimates are crucial. Previously however, mostly the past and present patient trajectory informed medical decisions. The estimation of the future patient trajectory is explored in detail in \autoref{figure2} and methodology is discussed in \autoref{section3}. With diagnostic information, uncertainty is reduced about differential diagnoses and the diagnostic net benefit of specific diagnostic actions estimated to facilitate diagnostic decisions leading to an action or diagnosis. Looping through this flowchart multiple times creates diagnostic paths which are shown in \autoref{figure1}B.
        \textbf{B) Diagnostic decision-making for an individual patient.} This is a stylized representation to anchor ideas. Diagnostic decisions are made at decision nodes factoring in expected costs (E\textsubscript{cost}) and expected benefits (E\textsubscript{benefit}) for the different possible actions. These are estimated considering the subsequent diagnostic path. Since diagnostic actions are not deterministic, the chance nodes show how the same action may lead to different outcomes, each associated with a patient-specific probability and confidence intervals. The blue path is the path taken. The black paths are possible counterfactual paths and the purple path is an example counterfactual path. The blue and purple boxes show how probabilities for different diagnoses, depicted in pie charts, are updated along the respective path with each action taken until a final diagnosis is reached.}
    \label{figure1}
\end{figure}

Health and disease are not binary states but rather make up a continuum, and while some diseases such as cancers are staged, their dynamic nature is overall not well understood. Individuals progress through personal \textbf{disease trajectories}\cite{Dobson2019-zk}\cite{Porsteinsson2021-pu}\cite{Wijsenbeek2022-vc} that clinicians aim to describe using \textbf{diagnoses}. Not unlike GPS systems that help pilots determine their current location for subsequent steering, \textbf{diagnoses} enable clinicians to roughly pinpoint the patient’s problem to facilitate further investigation or treatment. However, the diagnostic process itself can be difficult to navigate. \autoref{figure1}A anchors the subsequent arguments by highlighting the overall structure of the diagnostic process and the need for integrating observed and estimated information to facilitate \textbf{diagnostic decisions}. In this subsection, we outline a framework demonstrating how to achieve personalized \textbf{early} and \textbf{timely} \textbf{diagnosis}.

\medskip
The symptoms that patients present with can typically be caused by various underlying \textbf{differential diagnoses}.\cite{Jain2017-ik} Each \textbf{differential} \textbf{diagnosis} associated with a patient’s symptoms has a probability of being a true \textbf{diagnosis} within a differential diagnostic probability space.\cite{Aberegg2020-nv} These probabilities with their associated confidence intervals can be used to express \textbf{diagnostic uncertainty}\cite{Winkler1991-nf} which is a fundamental problem in medicine\cite{Hatch2016-ce}\cite{Kennedy2017-xs}.
\textbf{Diagnostic uncertainty} is usually highest in early disease stages when signs and symptoms are nonspecific and information is scarce. While uncertainty diminishes over time to reveal an increasingly streamlined \textbf{disease trajectory} with clearer symptoms enabling easier \textbf{diagnosis}, treatment is often more beneficial at earlier time points in pathogenesis\cite{Neal2015-ob} when \textbf{diagnostic uncertainty} is higher. Thus, waiting for signs and symptoms to become more specific translates to more information and reduced uncertainty while the risk increases that the window for effective intervention may pass or that the benefit of intervention may be smaller. The alternative extreme, performing a multitude of diagnostic tests early on, would involve potential patient harm through (unnecessary) procedures, additional financial costs, false positives and overdiagnosis. Since a \textbf{diagnosis} is rarely made with 100\% certainty, a balance needs to be struck between the two extremes allowing clinicians to facilitate treatment with an acceptable level of remaining \textbf{diagnostic uncertainty}. This level of acceptable uncertainty is not objectively defined, a problem outside the scope of this work.

\medskip
Sources of \textbf{diagnostic information}, which is key for reducing uncertainty\cite{Klir2006-ia}, include observed and estimated information. Observed information encompasses medical expertise, the past \textbf{disease trajectory}, i.e. medical history, and the present \textbf{disease trajectory} which is reflected in current signs and symptoms. If the observed information is inconclusive for finding a \textbf{diagnosis}, additional \textbf{diagnostic information} needs to be gathered through \textbf{diagnostic actions} that can produce test results or reveal new signs and symptoms. \textbf{Diagnostic actions} have associated \textbf{diagnostic considerations}, i.e. costs and benefits. Costs include the patient harm associated with performing \textbf{diagnostic actions} including those following possible misdiagnosis, while benefits of detecting early disease stages include less invasive treatment, longer survival, improved quality of life and cost-effectiveness for the healthcare system. \textbf{Diagnostic considerations} depend on the speed and nature of disease progression, the expected impact of actions taken in the process and patient preferences. Importantly, \textbf{diagnostic actions} are not deterministic and lead to various outcomes, each associated with a probability (see \autoref{figure1}B).

\medskip
To holistically estimate the considerations for a \textbf{diagnostic action} and thus facilitate \textbf{diagnostic decisions} among possible actions, clinicians need to take into account not only the patient’s unique past and present trajectory but also the expected future \textbf{disease trajectory} including the evolution of patient-specific information and disease progression over time. For instance, if the patient is projected to deteriorate significantly, a \textbf{diagnostic action} facilitating \textbf{early diagnosis} and treatment would offer increased benefits. Methods to facilitate such estimates become increasingly available (see \autoref{section3}).

\medskip
Often several \textbf{diagnostic actions} will be combined and form a \textbf{diagnostic path}. Typically, many potential \textbf{diagnostic paths} exist and thus represent counterfactual scenarios each associated with a \textbf{diagnostic net benefit} (see \autoref{figure1}B). Along the path, \textbf{diagnostic decisions} are made with the available \textbf{diagnostic information}. Those require consideration not only of the immediate next step but also a projection of the logical succession of actions contingent on the probabilities of outcomes from preceding actions. Importantly, as clinicians and patients move down the \textbf{diagnostic path}, each \textbf{diagnostic action} results in the acquisition of new information and thereby reduced \textbf{diagnostic uncertainty}. At decision nodes, the next \textbf{diagnostic decision} must be made, factoring in the newly available information and expectations about information acquired in subsequent steps (see \autoref{figure1}B). Therefore, decisions must be made dynamically throughout the \textbf{diagnostic path}.

\medskip
\textbf{Timely diagnosis} encapsulates the notion of an opportune or ‘right’ time for \textbf{diagnosis} aligning with contingent circumstances as opposed to simple chronological time.\cite{Van_der_Schaar2018-jt} To be useful, \textbf{timely diagnosis} depends on balancing individual \textbf{diagnostic considerations}\cite{Nuffield_Council_on_Bioethics2009-zm} which take effect over the course of the patient’s lifetime and may not be immediately apparent. The \textbf{diagnostic net benefits} of various \textbf{diagnostic paths} depend on the patient’s unique characteristics but also on the timing of actions, meaning that counterfactual\cite{Ginsberg1986-qq} scenarios need to be estimated to ultimately answer the question: What would be the consequences of performing a specific \textbf{diagnostic action} at a specific point in time for a specific patient?

\medskip
While previously impossible, this reasoning can be enabled by machine learning and statistics (see \autoref{section3}), which can 1) estimate the future patient trajectory and thereby complement observed information and 2) provide estimates of the \textbf{diagnostic net benefits} associated with counterfactual paths (see \autoref{figure2}). 

\medskip
We would like to mention that \textbf{timely} and \textbf{early diagnosis} are not opposing concepts. Depending on the nature of diagnostic tests, the expected benefit from early detection, patient preference etc., ‘timely’ can overlap with ‘very early’. \textbf{Timely diagnosis}, however, encapsulates a more comprehensive understanding of the intricate dynamics involved in the diagnostic process. Therefore, we will mostly refer to \textbf{timely diagnosis} in the following.

\subsection{An illustrative patient example}\label{subsec3}

To bring this framework to life, we present a simplified patient example from first presentation to receiving a \textbf{timely diagnosis}. We want to highlight that while we use a specific patient example to anchor ideas, the proposed framework is general and can be applied to every diagnosis process.

\medskip
A 62-year old woman, Susan, presents to her GP with persistent cough over 5 weeks, keeping her awake at night. The GP needs to establish a \textbf{diagnosis} for Susan’s cough to facilitate subsequent treatment.

\medskip
For brevity, in the case of Susan we will make the simplifying assumption that her cough is caused by only one of many \textbf{differential diagnosis}. The most common causes of cough in various populations are bronchitis (25 - 50\%), influenza (6-15\%) or asthma (3-15\%) while severe diseases such as pneumonia (4\%), COPD (1-3\%), and suspected malignancy (0.2-2\%) are rare.\cite{Bergmann2021-xg} Without further information the true \textbf{diagnosis} remains uncertain.

\medskip
To better estimate the cause of Susan’s cough and go beyond population-based data, additional \textbf{diagnostic information} contextualizes the presenting problem, has the capacity to reduce uncertainty\cite{Klir2006-ia} and offer more precise estimates of the probabilities for \textbf{differential diagnosis}. \textbf{Diagnostic information} observed from her EHR about her past medical history and information on her present signs and symptoms can be combined to adjust the probabilities of \textbf{differential diagnosis} with some becoming less likely (e.g., bronchitis, influenza, COPD, GERD) and others becoming more likely (e.g., ACE inhibitor side effect, bronchogenic carcinoma, post-infectious cough).

\medskip
Susan and her GP now have to make a \textbf{diagnostic decision} among various possible \textbf{diagnostic actions} each with specific \textbf{diagnostic considerations}. In Susan’s case, watch and wait may be a good approach if the \textbf{diagnosis} turns out to be post-infectious cough. However this would delay the revelation of other \textbf{diagnoses}. Similarly, stopping ACE inhibitor therapy could reveal whether side effects from the medication are the cause of Susan’s cough within 1-4 weeks although this may take up to 3 months.\cite{Dicpinigaitis2006-jz} Even though it is less likely, if Susan turns out to have lung cancer, a lot of time would have passed by then. This \textbf{differential} \textbf{diagnosis} could be confirmed quicker with tests that involve a higher risk for patient harm and are more time- and cost-consuming, such as CT scans and bronchoscopy. 

\medskip
To find the \textbf{diagnostic path} with the highest \textbf{diagnostic net benefit} to Susan, her GP needs to know more about the expected future \textbf{disease trajectory }including disease progression over time. This would allow her to understand and communicate the \textbf{net benefit} of every possible \textbf{diagnostic action} considering the subsequent counterfactual \textbf{diagnostic paths}.

\medskip
Given Susan’s unique characteristics and the observed and estimated \textbf{diagnostic information}, the \textbf{diagnostic path} with the highest \textbf{net benefit} may be to immediately perform chest X-ray which shows a lesion suggestive of non-small cell lung carcinoma (NSCLC). A subsequent \textbf{diagnostic decision} is made that leads to a CT scan and a CT guided biopsy that confirms the \textbf{diagnosis} of NSCLC. She and her GP have balanced the \textbf{diagnostic considerations}, saved time and reached a \textbf{timely diagnosis}.

\section{Methods}\label{section3}
\subsection{Machine Learning and Statistics to Enable Personalized Timely Diagnosis}\label{subsec3_1}

\begin{figure}
    \centering
    \includegraphics[width=1\textwidth]{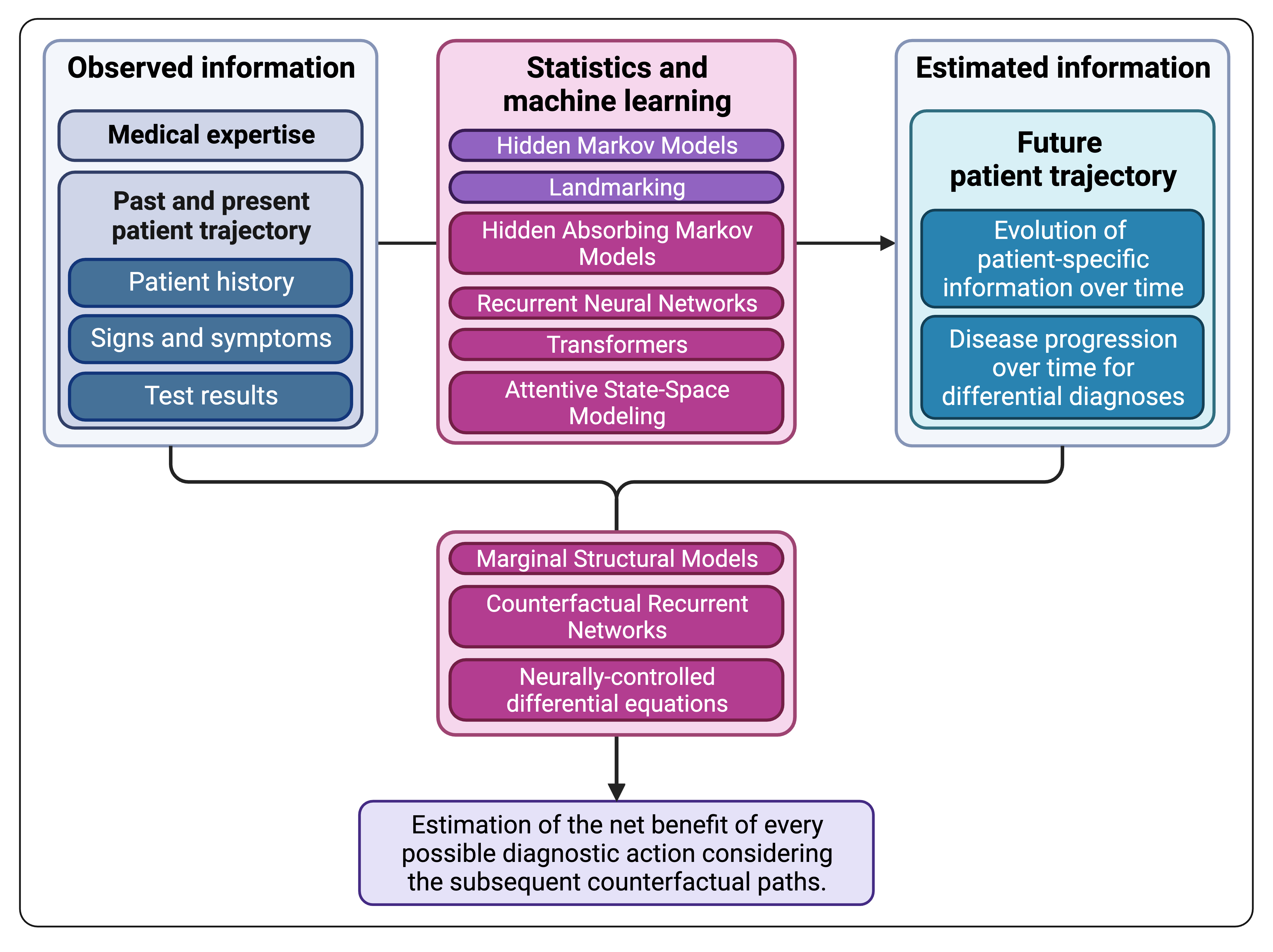}
    \caption{\textbf{Observed and estimated \textbf{diagnostic information} allow the calculation of \textbf{net benefits} of \textbf{diagnostic actions} for the individual patient.} To implement the timely diagnosis framework, the future patient trajectory needs to be estimated from observed information using statistics and machine learning methodology (red boxes). Subsequently, all available information can be leveraged to estimate the net benefit of potential diagnostic actions in the context of subsequent counterfactual paths. Methodology from statistics and machine learning to enable these estimates is discussed further in \autoref{section3}.
}
    \label{figure2}
\end{figure}

Currently, the nature of \textbf{diagnostic considerations} usually hinges on subjective estimates, rendering them susceptible to various heuristics and biases.\cite{Elstein2002-eh} The framework described above can guide the development and implementation of methods from statistics and machine learning to leverage the observed information and forecast personalized patient trajectories (see \autoref{figure2}).

\medskip
In this section, we will showcase methodology from statistics and machine learning that can combine medical expertise and observed information to 1) estimate the future patient trajectory to complement observed information and 2) assess the \textbf{net benefit} associated with alternative (counterfactual) \textbf{diagnostic paths}.

\medskip
Today, \textbf{diagnostic paths} have been developed and evaluated using population data to produce a single approach to be followed for all patients, albeit recognizing that individual patients may exit at different points. There may also be some stratification based on covariates such as age, gender, ethnicity, smoking status, alcohol consumption or other environmental factors.\cite{Hull2020-yl} This creation of increasingly more substrata to take into account multiple factors is perceived as moving toward a personalized pathway. However, the pathways for these substrata are still derived from population averages in terms of the impact of each covariate, generate variations on the same underlying pathway, and do not represent true personalization. Patients still follow a single population level diagnostic flowchart with adjustment for covariates that seem important at the population level. This is one flowchart for everyone and patients are stratified according to this one flowchart. In contrast, our framework characterizes and takes into account the impact of individual variability, not constrained by previously defined, population-level stratification factors, to create an individualized path where the timing and decision order are personalized to maximize the \textbf{diagnostic net benefit} for the individual patient.

\medskip
To make the framework actionable, the information available about the individual patient (i.e. the 'observed information') needs to be used to model the individualized future patient trajectory and estimate both the changing latent disease presentation over time and the associated, observed biomarker evolution. Subsequently, the observed and estimated information is combined to estimate the individualized benefits and costs of every counterfactual \textbf{diagnostic action} in the context of the entire \textbf{diagnostic path} (see \autoref{figure2}).

\medskip
To model individualized patient trajectories from the observed data, thereby maximizing available \textbf{diagnostic information}, statistical methods such as Hidden Markov Models\cite{Sukkar2012-bb} or Landmarking\cite{Putter2022-ai}, or machine learning methods such as Hidden Absorbing Markov Models\cite{Alaa2018-dx}, Recurrent Neural Networks\cite{Wang2018-uq}, Transformers\cite{Nguyen2023-ng} or Attentive State-Space Modeling\cite{Alaa2019-jp} can be used. 
To estimate counterfactual outcomes over time and ultimately the \textbf{net benefit} of every possible \textbf{diagnostic action}, statistical or machine learning approaches like Marginal Structural Models\cite{Williamson2017-pp}, Counterfactual Recurrent Networks\cite{Bica2020-hg} or neurally-controlled differential equations\cite{Seedat2022-ip} can be applied.

\medskip
These models can be trained on complex multidimensional datasets from large patient cohorts and then accessed and used in real time to output personalized \textbf{disease trajectories} and counterfactual estimates of the \textbf{net benefits} of possible \textbf{diagnostic paths} for the individual patient consulting a clinician. Some of these models will also be able to provide uncertainty estimates associated with the predictions and counterfactuals as well as offer explainability in terms of important features over time (including changes in features), similar patients, etc. Finally, as new information becomes available about the patient, these models can continuously improve their forecast accuracy and relevance for the individual patient.

\subsection{Implementation through a Decision Support Tool}\label{subsec3_2}
To empower informed decision-making, data-driven predictions from statistics or machine learning models can provide crucial insights to patients and clinicians through individualized, responsive decision support tools.
Several prerequisites are needed to enable their implementation. 

\medskip
First, to make personalization possible, the framework requires a detailed mapping of the diagnostic process and, on the basis of this mapping, collection of necessary patient data (e.g. EHRs, genetic information, lifestyle factors), and dynamic integration of newly acquired patient information to refine predictive capabilities. Gathering this information will require patient engagement, informed consent, and safeguards to comply with privacy standards.

\medskip
Second, user-friendly interfaces will need to be developed that seamlessly call the necessary models in the background and integrate with IT systems that clinicians are already familiar with. Existing decision support tools show that integration with EHRs is feasible.\cite{Tillotson2017-qc}\cite{Frymoyer2020-vg}
Interfaces should provide individualized estimates of the future \textbf{disease trajectory}, \textbf{diagnostic considerations} associated with possible \textbf{diagnostic actions} and \textbf{paths}, and uncertainty estimates. Since shared decision-making is becoming an integral part of practicing medicine, patient-specific interfaces should also be provided to offer information on potential options in an easily accessible format.

\medskip
Finally, decision support tools developed with this framework need to undergo clinical testing and validation to ensure their efficacy and safety. Moreover, clinicians need to be empowered with the skills to effectively and safely use these tools and interpret their outputs.

\section{Final Remarks}\label{section4}
While in this paper we mainly focused on patient outcomes and highlighted how the personalized framework for \textbf{timely diagnosis} supports clinicians and patients, it can also be used to positively impact entire healthcare systems. Using this foundational framework, scarce resources can be allocated effectively. For instance, the value of time- and capital-intensive tests such as MRI scans or emerging tests such as liquid biopsies for cancer-detection can be assessed for individual patients. Thus, there is potential for improving patient outcomes while reducing overall cost. The framework scales to different types of healthcare systems and could particularly benefit LMIC countries.

\medskip
We acknowledge that our proposal is an ambitious project requiring interdisciplinary collaboration among all stakeholders including patients, clinicians, data scientists, statisticians, machine learners, politicians, and ethicists. That being said, we believe that the proposed foundational framework can be implemented and that substantial effort is justified given the extraordinarily high stakes involved in \textbf{early diagnosis}.

\medskip
We hope for this foundational framework to serve as a catalyst for the implementation of clinically impactful decision support tools and conscious allocation of diagnostic tests that will usher in a future where \textbf{timely diagnosis} and treatment improve the health of each individual patient.
\bigskip

\backmatter

\bmhead{Acknowledgments}
We would like to thank Andrew Rashbass, Eoin McKinney, Tom Callender, Henk van Weert, Ari Ercole, Qiyao Wei, Johann Maaß, Christian P Schaaf, Ferdinand Althammer, Tim Oosterlinck, Jesse Maile, Moritz Flotho, and the participants of the 29th Revolutionizing Healthcare engagement session for our insightful discussions and Quirin Krabichler for his help in creating figures. TS receives scholarships from the German Academic Scholarship Foundation (Studienstiftung des deutschen Volkes) and the Medical Faculty of Heidelberg University, Germany. Figures were created with Biorender.com.

\bmhead{Ethics Declarations}
TS, AG and CD declare no conflicts of interest. 
RWP holds stock in and receives compensation from F Hoffmann la Roche.
MvdS leads the Cambridge Centre for AI in Medicine; this center has received research funding from AstraZeneca and GlaxoSmithKline.

\newpage
\bibliography{sn-article.bib}

\newpage
\begin{appendices}

\section{Screening and Early Diagnosis}\label{appendix}

The WHO postulates early diagnosis only for individuals with symptoms.\cite{World_Health_Organization2017-jh} However, a pathology can be present, and patients can be on a disease trajectory even before symptoms occur. This is not purely a matter of risk when pathogenesis is already underway. This earlier stage could, for example, be preinvasive cancer. Whether one would reach a ‘diagnosis’ when determining that a patient is on a disease trajectory at an early stage or whether one would find a higher risk for developing invasive cancer depends on whether preinvasive cancer is defined as an entity in the ICD.

\medskip
However, the framework proposed here does not require symptoms to be present. It only requires a single hint that a person is not perfectly healthy to get started. Of course, the less clear signs are, the less likely diagnostic testing will have a sufficiently high net benefit to be considered as an option as opposed to waiting.

\medskip
The WHO defines the difference between screening and early diagnosis mainly according to symptom onset.\cite{World_Health_Organization2017-jh} However, there is a fundamental difference of at least equal importance: screening targets a population, and early diagnosis targets individuals. This framework would inform something akin to ‘screening’ for a specific person, which may instead be called early detection (as to keep the distinction from symptomatic patients). This framework can just as well inform users about the utility of different tests in presymptomatic contexts.

\end{appendices}

\end{document}